\documentclass[table]{article} 
\usepackage{iclr2025_conference,times}

\usepackage{amsmath,amsfonts,bm}

\def\eqref#1{equation~\ref{#1}}

\def\1{\bm{1}}

\DeclareMathAlphabet{\mathsfit}{\encodingdefault}{\sfdefault}{m}{sl}
\SetMathAlphabet{\mathsfit}{bold}{\encodingdefault}{\sfdefault}{bx}{n}

\usepackage{hyperref}
\usepackage{fontawesome}
\usepackage{graphicx}
\usepackage{booktabs}
\usepackage{caption}
\usepackage{array}
\usepackage{wrapfig}
\usepackage{url}
\usepackage{xcolor}
\usepackage{siunitx} 
\usepackage{subcaption}  

\definecolor{lightblue}{RGB}{230, 245, 255}
\definecolor{lightred}{RGB}{255, 230, 230}

\iclrfinalcopy

\title{ENA: Efficient N-dimensional Attention}

\author{Yibo Zhong \\
Independent Researcher \\
\texttt{yibozhong657@gmail.com} \\
}

\begin{document}

\maketitle

\begin{abstract}
\textit{\textbf{TL;DR:}} \textit{A layer-interleaved hybrid architecture of linear recurrence and attention (full or local) matches Transformer performance on high-order data with greater efficiency.}

Efficient modeling of long sequences of high-order data requires a more efficient architecture than Transformer. In this paper, we investigate two key aspects of extending linear recurrent models, especially those originally designed for language modeling, to high-order data (1D to ND): scanning strategies and attention-hybrid architectures. Empirical results suggest that scanning provides limited benefits, while attention-hybrid models yield promising results. Focusing on the latter, we further evaluate types of attention and find that tiled high-order sliding window attention (SWA) is efficient in both theory and practice. We term the resulting hybrid architecture of linear recurrence and high-order SWA as Efficient N-dimensional Attention (ENA). We then conduct several experiments to demonstrate its effectiveness. The intuition behind ENA is that linear recurrence compresses global information into a state, while SWA complements it by enforcing strict local modeling. Together, they form a simple framework that offers a promising and practical solution for ultra-long high-order data modeling. \quad \\
  \href{https://github.com/fla-org/fla-zoo}{\faGithub~Models}\ \quad
  \href{https://github.com/lululu39/REPA-Plus}{\faGithub~2D Gen Training}\ \quad
  \href{https://github.com/lululu39/ena}{\faGithub~Training Code}\ \quad
  \href{https://huggingface.co/fla-zoo}{
    \includegraphics[height=1em]{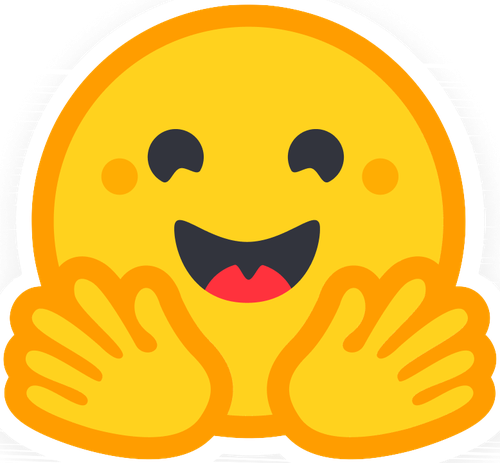}~Weights
  }
\end{abstract}

\section{Introduction}

\begin{wrapfigure}{r}{0.44\textwidth}
\centering
\vspace{-10pt}
\includegraphics[width=0.40\textwidth]{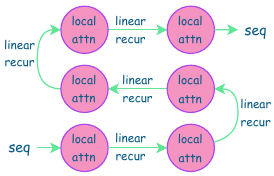}
\caption{An overview of the architecture of ENA with alternating layers of linear recurrence and local attention. Note that although we ultimately perform no sequence permutation, the framework remains compatible with any scanning.}
\vspace{-15pt}
\label{fig:arc}
\end{wrapfigure}

Softmax attention in LLMs has quadratic time complexity, making it inefficient for long sequences. To address this, linear recurrent models\footnote{In this paper, "linear recurrent models", "linear recurrence" and "linear models" are used interchangeably to denote models that perform sequence modeling via state updates with linear time complexity. The same applies to "softmax attention" and "full attention". } have emerged as efficient alternatives. Representative variants include RetNet \citet{sun2023retnet}, HGRN \citet{qin2024hgrn2}, GLA \citet{yang2024gla}, GSA \cite{zhang2024gsa}, Mamba \cite{gu2023mamba} and RWKV \citet{peng2023rwkv}. Subsequent advancements such as DeltaNet \citet{yang2024deltanet}, Gated DeltaNet \citet{yang2024gdn}, DeltaProduct \cite{siems2025deltaP}, LaCT \cite{zhang2025tttdoneright} and MesaNet \cite{vonoswald2025mesanet} further enhance expressiveness while preserving linear-time complexity and are optimized for parallel training.

\textbf{Bridging the dimensional gap with scanning}

\begin{figure}[t]
    \centering
     \vspace{-30pt}
    \includegraphics[width=0.65\linewidth]{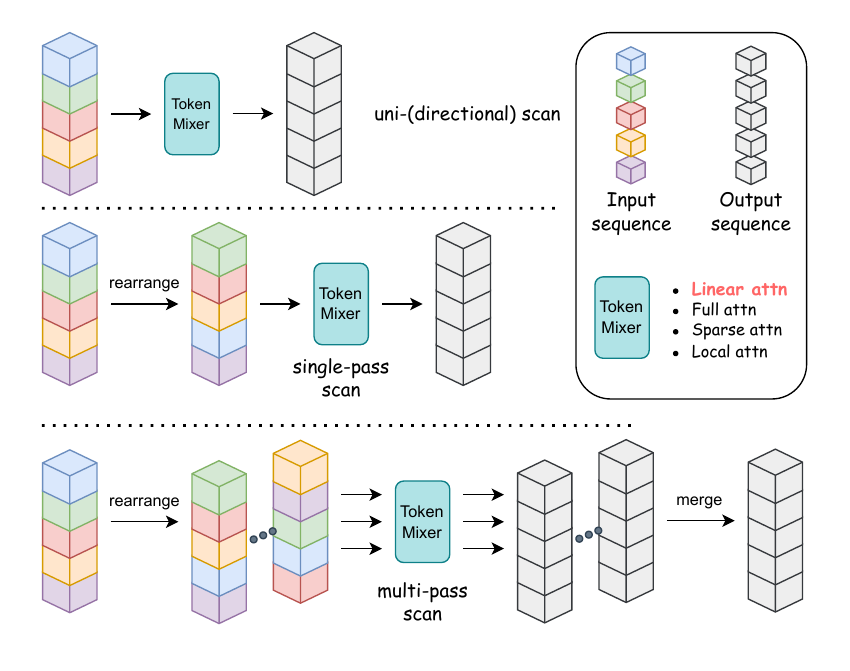}
    \vspace{5pt}
    \caption{A simple illustration of the operations performed by different scanning methods. Single-pass scan rearranges the input into one sequence, while multi-pass scanning generates multiple input sequences. For multi-pass scanning, the output sequences are merged into a single output sequence. Uni-directional scan is a special case of single-pass scan where no rearrangement is performed. The rules for rearrangement and merging are mostly predefined and fixed.}
    \vspace{-5pt}
    \label{fig:scan-catagory}
\end{figure}

Drawing inspiration from their efficacy in language modeling, these linear models have been extended to high-order data modeling \footnote{For convenience, we refer to all data with more than one dimension as higher-order data, including images, which have only two dimensions.}. To reconcile the inherent 1-dimensional nature of linear models with the N-dimensional structure of high-order data (e.g., images, videos), prior works typically employ various "scanning" methods. These methods transform an input into single (single-pass scan) or multiple (multi-pass scan) derived sequences using predefined patterns, subsequently processed by a shared linear model, as illustrated in Fig. \ref{fig:scan-catagory}.

However, multi-pass scanning methods process multiple sequences at once (serially or in parallel), causing significant speed and memory overhead. Consequently, in a simple stacked-layer architecture, multi-pass scanning often renders linear models even slower than a Transformer using FlashAttention \citep{dao2023flashattention2}. To address this inefficiency, many works introduce complex architectures such as downsampling to reduce computational cost. Recent works \citep{wang2024adventurer, zhu2024dig} propose single-pass approaches that permute the sequence between blocks while ensuring only a single sequence is fed to the linear model. However, the effectiveness of these single-pass methods has not yet been comprehensively and fairly evaluated.
\begin{figure}[t]
    \centering
    \vspace{-30pt}
    \includegraphics[width=0.95\linewidth]{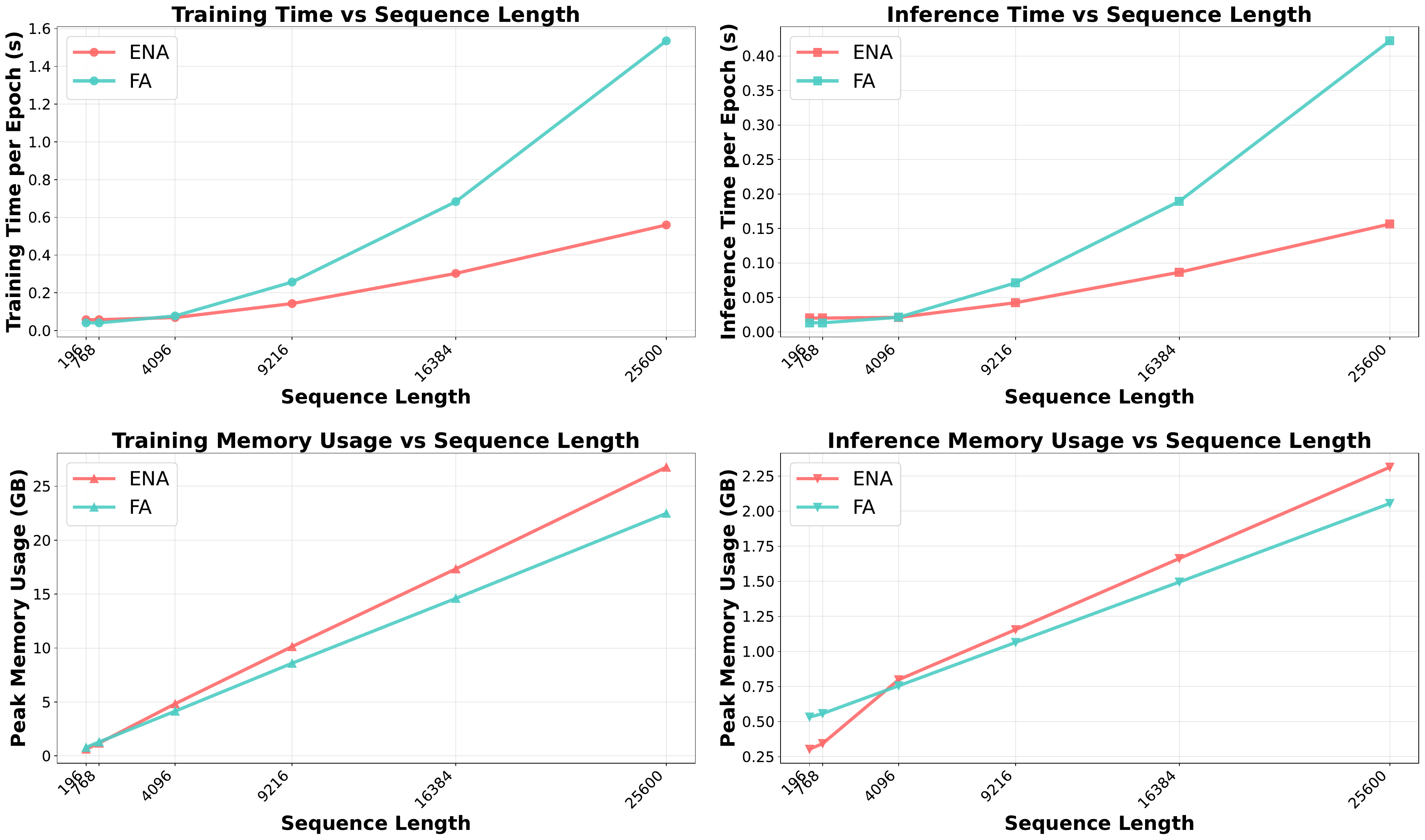}
    \vspace{0pt}
    \caption{Performance comparison between ENA and Flash Attention (FA)-based Transformer vision encoders across different sequence lengths. Training and inference times (top row), as well as peak memory usage (bottom row), are measured with a batch size of 4 over 50 epochs using mixed precision on a single GPU. For sequence lengths $\geq 1152$, ENA employs 2D STA (tile size: $16 \times 8$) with a window size covering approximately $30\%$ of the tokens, resulting in $70\%$ sparsity.}
    \vspace{-25pt}
    \label{fig:speed}
\end{figure}

\textbf{Bridging the dimensional gap with attention-hybrid architectures instead}

Alternatively, hybrid architectures bridge the dimensional gap by combining linear recurrence with attention. While linear recurrence compresses information into a fixed-size state, it may overlook important local patterns. Introducing attention helps compensate for this limitation. For long sequences, we primarily consider local attention, where attention is computed within a fixed-size neighborhood. This design leverages the inherent locality in data while maintaining computational efficiency. For short sequences, we adopt full attention in the hybrid architecture.

We categorize local attention for high-order data into block attention and sliding window attention (SWA). Block attention divides the sequence into non-overlapping blocks and computes attention within each block. In contrast, SWA defines local regions centered around each token (or token tile in sliding tile attention), enabling finer-grained locality modeling. Our experiments show that a hybrid architecture using high-order SWA outperforms one using block attention.

Based on our findings, we propose \textbf{Efficient N-dimensional Attention (ENA)}, a simple hybrid architecture that combines linear recurrence and high-order SWA, as illustrated in Fig. \ref{fig:arc}. ENA maintains a straightforward architecture of stacked layers and employs a simple block-wise hybrid strategy for ease of implementation, thus avoiding unnecessary complexity. For ENA, we primarily use DeltaNet~\cite{yang2024deltanet} with no sequence permutation (i.e., scanning) as the linear model. For hardware-efficient high-order SWA, we use Sliding Tile Attention (STA \cite{zhang2025sta_attn}).

This paper is organized as follows: we first introduce two ways for bridging the dimensional gap: scanning and integrating attentions, accompanied by a small-scale pilot experiment. We then evaluate various design choices via ImageNet classification. After establishing ENA's architecture, we conduct several experiments to validate its effectiveness. Finally, we provide additional discussions.

\begin{figure}[t]
    \centering
    \vspace{-20pt}
    \includegraphics[width=0.99\linewidth]{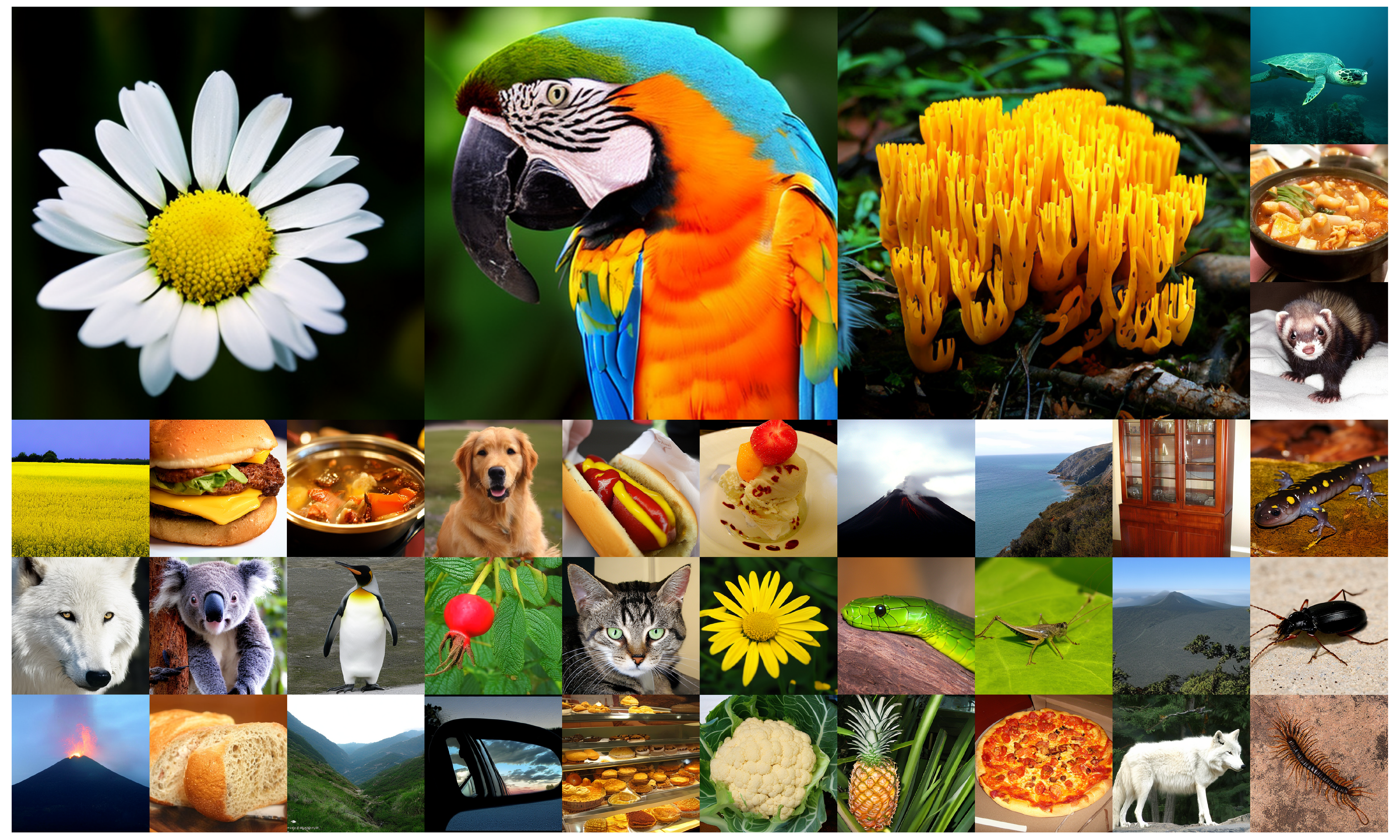}
    \caption{Selected image generation results from \textit{ena-deltanet-sta-w24x24-t8x8-xl-gen2d} on ImageNet with a resolution of 512 $\times$ 512.}
    \label{fig:gen2d}
    \vspace{-0pt}
\end{figure}

\section{From 1D to ND}

This section introduces methods for adapting the 1D nature of linear models to high-order data. We discuss two main approaches: scanning methods for pure linear models and attention-based hybrid architectures.

\begin{table*}[t]
\vspace{-0pt}
\centering
\caption{Overview of different scanning methods considered. $B$ represents batch size, $L$ represents sequence length, and $D$ represents the feature dimension. In multi-pass methods, we arrange multiple sequences along the batch dimension to enable parallel processing, even though prior works might perform these operations sequentially. Since single-pass scanning is underexplored, we introduce several new scan types for comparison, marked in light blue and light red. Methods highlighted in light red leverage the multi-head design in token mixers to enable multi-directional scan within a single pass.}
\vspace{-5pt}
\label{tab:scan-types}
\resizebox{0.98\textwidth}{!}{%
\begin{tabular}{@{}l p{1.0cm} p{12.0cm}@{}}
\toprule
\textbf{Scan Type} & \textbf{Pass} & \textbf{Operation Flow} \\
\midrule
\textbf{uni-scan} & Single & \texttt{[B,L,D] → token mixer → [B,L,D]} \\
\textbf{switch-scan} & Single & \texttt{[B,L,D] → token mixer → flip/transpose → [B,L,D]} \\
\textbf{flip-scan} & Single & \texttt{[B,L,D] → flip → token mixer → [B,L,D]} \\
\rowcolor{lightblue}
\textbf{1d-shift-scan} & Single & \texttt{[B,L,D] → token mixer → shift → [B,L,D]} \\
\rowcolor{lightblue}
\textbf{2d-shift-scan} & Single & \texttt{[B,L,D] → token mixer → reshape → 2D shift → [B,L,D]} \\
\rowcolor{lightblue}
\textbf{random-scan} & Single & \texttt{[B,L,D] → random shuffle → token mixer → [B,L,D]} \\
\rowcolor{lightblue}
\textbf{learnable-scan} & Single & \texttt{[B,L,D] → learnable module → token mixer → [B,L,D]} \\
\rowcolor{lightred}
\textbf{multi-head-bi-scan} & Single & \texttt{[B,L,D] → head-wise flip → token mixer → head-wise reverse flip → [B,L,D]} \\
\rowcolor{lightred}
\textbf{multi-head-2d-scan} & Single & \texttt{[B,L,D] → head-wise rearrange → token mixer → head-wise reverse rearrange → [B,L,D]} \\
\midrule
\textbf{bi-scan} & Multi & \texttt{[B,L,D] → flip → [2B,L,D] → token mixer → [2B,L,D] → merge → [B,L,D]} \\
\textbf{cross-scan} & Multi & \texttt{[B,L,D] → rearrange → [4B,L,D] → token mixer → [4B,L,D] → merge → [B,L,D]} \\
\bottomrule
\end{tabular}%
}
\vspace{-7pt}
\end{table*}

\subsection{Scanning for Linear Recurrence}

The diverse scanning methods outlined in Table~\ref{tab:scan-types} can be described with a general formulation. Let $\mathbf{X} \in \mathbb{R}^{B \times L \times D}$ be the input tensor, where $B$ is the batch size, $L$ is the sequence length, and $D$ is the feature dimension. The core of each method involves a token mixer operation, denoted as $\mathcal{TM}(\cdot)$, which maps an input tensor to an output tensor of the same shape.

For single-pass methods, the transformation can be expressed generally. For example, the \textbf{uni-scan} performs a direct token mixing:
\begin{equation}
\mathbf{Y} = \mathcal{TM}(\mathbf{X})
\end{equation}
Methods that permute the sequence such as \textbf{flip-scan} and \textbf{switch-scan} can be unified as applying a pre-processing mapping $OP_{pre}(\cdot)$ and a post-processing mapping $OP_{post}(\cdot)$ around the token mixer:
\begin{equation}
\mathbf{Y} = OP_{post}(\mathcal{TM}(OP_{pre}(\mathbf{X})))
\end{equation}
Here, $OP_{pre}(\cdot)$ and $OP_{post}(\cdot)$ may be identity mappings or concrete operations such as $flip(\cdot)$ or $shift(\cdot)$, depending on the specific method.

For multi-pass methods like \textbf{bi-scan}, the input $\mathbf{X}$ is first transformed into multiple ($N$) views (e.g., $\mathbf{X}_1 = \mathbf{X}$, $\mathbf{X}_2 = \mathrm{flip}(\mathbf{X})$ for $N=2$). These views are processed in parallel using a shared token mixer $\mathcal{TM}(\cdot)$, and the outputs are merged into a single sequence:
\begin{equation}
\mathbf{Y} = \frac{1}{N} \sum_{i=1}^{N} \mathcal{TM}(\mathbf{X}_i)
\end{equation}
The \textbf{cross-scan} follows a similar pattern with $N=4$ views. Note that the merge operation can be more sophisticated than simple summation; however, we adopt summation for simplicity.

While we include two common multi-pass methods, our primary focus is on the less explored single-pass approaches. To facilitate a comprehensive comparison, we introduce several single-pass variants: \textit{1D scan} and \textit{2D shift-scan}, which shift tokens along spatial or flattened dimensions, respectively; \textit{random-scan}, which applies a random token permutation prior to the token mixer; and \textit{learnable-scan}, which learns token reordering through a trainable permutation matrix, employing Gumbel-Softmax for differentiable selection. \textit{Learnable-scan} is initialized near identity and trained end-to-end. However, it is inefficient and, in our current implementation, can lead to the omission of tokens due to repeated selections. Finally, we also introduce multi-head scanning methods, including \textit{multi-head-bi-scan} and \textit{multi-head-2d-scan}, which divide the attention heads into groups and assign each group a dedicated scanning direction. These methods aim to achieve multi-directional scanning within a single pass; however, \textbf{in our current experiments, they yield no observable performance gains over a simple uni-scan in our experiments.}.

\subsection{Integrating Local Attention}

\label{local_attn_def}

An alternative approach incorporates auxiliary token mixers to compensate for inherent limitations of linear models, such as their lack of fine-grained local information. The linear model compresses global context, while the auxiliary mixer captures local patterns, creating a complementary architecture. A natural consideration for such token mixers is local attention, in which attention is restricted to a fixed window, leveraging the inherent locality of the data.

The block-wise hybrid approach used in \ref{fig:arc} can be formulated as follows. Let $\mathbf{X}^{(0)}$ denote the initial embeddings. For the $i$-th block in a model, the output $\mathbf{X}^{(i)}$ is computed as:
\begin{equation}
\begin{aligned}
\mathbf{X}'^{(i)} &= \mathbf{X}^{(i-1)} + \mathcal{TM}^{(i)}(\mathcal{N}(\mathbf{X}^{(i-1)})) \\
\mathbf{X}^{(i)} &= \mathbf{X}'^{(i)} + \mathcal{CM}(\mathcal{N}(\mathbf{X}'^{(i)}))
\end{aligned}
\end{equation}
where $\mathcal{N}$ denotes normalization and $\mathcal{CM}$ is the channel mixer. The token mixer $\mathcal{TM}^{(i)}$ alternates according to the block index:
\begin{equation}
\mathcal{TM}^{(i)} =
\begin{cases}
\mathcal{TM}_{\text{linear}} & \text{if } i \bmod 2 = 1 \\
\mathcal{TM}_{\text{local}} & \text{if } i \bmod 2 = 0
\end{cases}
\end{equation}
Here, $\mathcal{TM}_{\text{linear}}$ denotes the linear model, and $\mathcal{TM}_{\text{local}}$ denotes local attention. This alternating strategy aims to balance global context aggregation with local feature refinement.

We categorize local attention into two main types:

\begin{enumerate}
    \item Local attention with non-overlapping blocks (Block attention). This type of local attention is easy to implement with Flash Attention and achieves high hardware utilization. However, as we will demonstrate in our experiments, this simple partitioning impose restrictions on queries near the block borders, resulting in sub-optimal performance. Swin~\cite{liu2021swin} addresses this by shifting blocks across layers, but at the cost of increased design complexity.
    
    \item High-order sliding window attention (SWA). These types of local attention are natural extensions of SWA from language modeling to higher-order domains, imposing no restrictions on border queries. However, as illustrated in~\cite{zhang2025sta_attn}, naive implementations~\citep{hassani2023natten, liu2024clear} suffer from poor hardware utilization and thus offer little to no speedup over full attention, despite having significantly fewer FLOPs due to high attention sparsity.
\end{enumerate}

As prior works point out, the primary reason for slow high-order SWA is the generation of mixed blocks in the attention map, where only some elements require computation. Therefore, Sliding Tile Attention (STA~\cite{zhang2025sta_attn}) addresses this by sliding the window tile by tile, which eliminates mixed blocks and achieves true hardware speedup over full attention. At a high level, STA is a coarser-grained variant of high-order SWA. This design choice enables it to leverage the underlying hardware efficiently and achieve tangible speedups.

\subsection{Pilot Experiments}

We conduct pilot experiments using tiny models on small-scale datasets (CIFAR-10/100 and Tiny ImageNet) to evaluate the effectiveness of multi-pass scanning methods (bi-scan and cross-scan), which are otherwise too slow to test at larger scales. While such methods aim to bridge the dimensional gap, they introduce additional computational and memory overhead. Our central question is: \textit{Is the extra cost of multi-pass scanning worth it?}

We compare bi-scan and cross-scan against uni-scan (baseline), random-scan (which helps prevent overfitting), and hybrids that insert a full attention layer at the second block. Linear models include DeltaNet~\cite{yang2024deltanet}, Gated DeltaNet~\cite{yang2024gdn}, GSA~\cite{zhang2024gsa}, HGRN, HGRN2~\cite{qin2024hgrn2}, and RetNet~\cite{sun2023retnet}. All models use 6 layers and under 10M parameters, trained for 50/25/30 epochs on CIFAR-100/10 and Tiny ImageNet, respectively, with a learning rate of $1 \times 10^{-4}$ and a 0.2 warm-up ratio.

As shown in Table~\ref{tab:pilot}, multi-pass scans achieve the best results in only 5 out of 18 settings, despite increased latency and memory usage. In contrast, inserting a single full attention layer yields consistent gains. These results demonstrate that the cost of multi-pass scanning is unjustified even in small-scale settings.

\begin{table*}[t]
    \centering
    \vspace{-0pt}
    \caption{Pilot experiments conducted with tiny models on several small datasets: CIFAR-100 (C100), CIFAR-10 (C10), and TinyImageNet (TIN). The terms \textit{rand} and \textit{attn1} denote random-scan and the addition of a full attention layer at the second block, respectively.}

    \begin{subtable}[t]{0.49\textwidth}
        \centering
        \caption{DeltaNet}
        \begin{tabular}{lccc}
            \toprule
            Method & C100 & C10 & TIN \\ \midrule
            uni & 16.66 & 60.30 & 19.66 \\
            rand & 27.22 & 62.93 & 28.62 \\
            bi & 20.91 & 59.82 & 20.94 \\
            cross & 22.58 & 61.50 & 22.20 \\
            uni + attn1 & 20.98 & 61.87 & 20.35 \\
            rand + attn1 & 29.17 & 63.90 & 28.82 \\
            \bottomrule
        \end{tabular}
    \end{subtable}
    \begin{subtable}[t]{0.49\textwidth}
        \centering
        \caption{GDN}
        \begin{tabular}{lccc}
            \toprule
            Method & C100 & C10 & TIN \\ \midrule
            uni & 24.12 & 68.47 & 20.35 \\
            rand & 35.00 & 70.42 & 28.22 \\
            bi & 28.81 & 66.75 & 21.61 \\
            cross & 31.95 & 69.14 & 26.16 \\
            uni + attn1 & 28.59 & - & 21.67 \\
            rand + attn1 & 35.22 & 70.85 & 28.11 \\
            \bottomrule
        \end{tabular}
    \end{subtable}

    \vspace{1em}

    \begin{subtable}[t]{0.49\textwidth}
        \centering
        \caption{GSA}
        \begin{tabular}{lccc}
            \toprule
            Method & C100 & C10 & TIN \\ \midrule
            uni & 21.96 & 67.18 & 25.01 \\
            rand & 24.29 & 58.02 & 27.71 \\
            bi & 26.31 & 63.02 & 26.29 \\
            cross & 27.43 & 62.20 & 24.92 \\
            uni + attn1 & 22.44 & 66.88 & 26.98 \\
            rand + attn1 & 23.35 & 58.22 & 26.39 \\
            \bottomrule
        \end{tabular}
    \end{subtable}
    \begin{subtable}[t]{0.49\textwidth}
        \centering
        \caption{HGRN}
        \begin{tabular}{lccc}
            \toprule
            Method & C100 & C10 & TIN \\ \midrule
            uni & 23.72 & 74.60 & 26.83 \\
            rand & 31.16 & 66.30 & 27.95 \\
            bi & 30.99 & 71.84 & 27.54 \\
            cross & 31.53 & 73.78 & 29.72 \\
            uni + attn1 & 37.32 & 75.35 & 29.01 \\
            rand + attn1 & 34.15 & 67.54 & 29.05 \\
            \bottomrule
        \end{tabular}
    \end{subtable}

    \vspace{1em}

    \begin{subtable}[t]{0.49\textwidth}
        \centering
        \caption{HGRN2}
        \begin{tabular}{lccc}
            \toprule
            Method & C100 & C10 & TIN \\ \midrule
            uni & 24.42 & 74.03 & 27.46 \\
            rand & 32.08 & 67.43 & 26.33 \\
            bi & 31.94 & 70.20 & 26.89 \\
            cross & 34.22 & 73.69 & 28.07 \\
            uni + attn1 & 34.28 & 75.34 & 27.73 \\
            rand + attn1 & 31.07 & 67.79 & 26.55 \\
            \bottomrule
        \end{tabular}
    \end{subtable}
    \begin{subtable}[t]{0.49\textwidth}
        \centering
        \caption{RetNet}
        \begin{tabular}{lccc}
            \toprule
            Method & C100 & C10 & TIN \\ \midrule
            uni & 26.60 & 70.65 & 25.53 \\
            rand & 33.91 & 69.56 & 28.82 \\
            bi & 35.26 & 69.05 & 26.10 \\
            cross & 37.72 & 70.34 & 29.25 \\
            uni + attn1 & 33.60 & 70.79 & 26.08 \\
            rand + attn1 & 33.66 & 69.09 & 27.46 \\
            \bottomrule
        \end{tabular}
    \end{subtable}
    \label{tab:pilot}
\vspace{0pt}
\end{table*}

\begin{table}[t]
\centering
\vspace{-0pt}
\caption{Top-1 accuracy of DeltaNet (30M parameters, 12 layers) on ImageNet-1K validation set. The input resolution is set to 224 with a patch size of 16, resulting in a sequence length of 196. The model is trained from scratch under various settings for 100 epochs, with a batch size of 2048 (unless otherwise specified, 2048 is used as the default batch size for ImageNet training). For comparison, a baseline transformer trained using the same codebase is also included.}
\
\renewcommand{\arraystretch}{1.05} 

\begin{tabular}{@{} >{\raggedright\arraybackslash}p{0.6\textwidth} c @{}} 
\toprule
Variant & Top-1 Acc (\%) \\ 
\midrule
\rowcolor{lightblue}
\multicolumn{2}{@{}l}{\textit{Baseline}} \\ 
\addlinespace
Transformers & 78.51 \\
\addlinespace
\midrule
\rowcolor{lightblue}
\multicolumn{2}{@{}l}{\textit{Single-pass Scan Variants}} \\ 
\addlinespace
uni-scan & 75.75 \\
random-scan & 68.91 \\
flip-scan & 72.78 \\
switch-scan & 72.63 \\
2dshift-scan & 72.16 \\
learnable-scan & 72.64 \\
\addlinespace
\midrule
\rowcolor{lightblue}
\multicolumn{2}{@{}l}{\textit{Multi-pass Scan Variants}} \\ 
\addlinespace
bi-scan & 75.45 \\
\addlinespace
\midrule
\rowcolor{lightblue}
\multicolumn{2}{@{}l}{\textit{Multi-Head Scan Variants}} \\
\addlinespace
multi-head-cross-scan & 75.34 \\
multi-head-bi-scan & 75.04 \\
\addlinespace
\midrule
\rowcolor{lightblue}
\multicolumn{2}{@{}l}{\textit{Uni-Scan with Full Attention Layers}} \\
\addlinespace
\hspace{1em}Layers: \textit{5,11} & 76.77 \\
\hspace{1em}Layers: \textit{0,6} & 77.20 \\
\hspace{1em}Layers: \textit{3,7,11} & 77.38 \\
\hspace{1em}Layers: \textit{0,2,4,6,8,10} & 78.28 \\
\hspace{1em}Layers: \textit{6,7,8,9,10,11} & 77.75 \\
\addlinespace
\midrule
\rowcolor{lightblue}
\multicolumn{2}{@{}l}{\textit{Uni-Scan with 1D Sliding Window Attention}} \\
\addlinespace
\hspace{1em}Layers: \textit{0,2,4,6,8,10}; Window Size: 32 & 77.16 \\
\addlinespace
\bottomrule
\end{tabular}
\label{tab:deltanet_imgnet1k_acc}
\end{table}

\begin{table}[htbp]
\centering
\caption{Validating the effectiveness of cross-scan. We report Top-1 accuracy (\%) on ImageNet-1K validation set. All models are trained from scratch for 100 epochs at a resolution of 224 with a patch size of 16, resulting in a sequence length of 196. Since cross-scan requires more GPU memory, we use a batch size of 1024. Uni-scan + full attn \{0, 6\} denotes hybrid architecture where layer 0 and 6 are replaced with full attention layers. We can observe that simply integrating two full attention layer could yield better result than cross-scan.}
\label{tab:cross_scan_compare}
\setlength{\tabcolsep}{5pt} 
\begin{tabular}{@{} >{\raggedright\arraybackslash}p{0.35\textwidth} S[table-format=2.2] @{}}
\toprule
Model / Configuration & {Top-1 Acc (\%)} \\
\midrule
uni-scan & 76.42 \\
cross-scan & 77.30 \\
uni-scan + full attn \{0, 6\} & 77.71 \\
\bottomrule
\end{tabular}
\end{table}

\begin{table}[htbp]
\centering
\caption{Top-1 accuracy (\%) of Single-pass Scan Variant using various linear model on ImageNet-1K validation set. We evaluate various scan methods (uni-scan, flip-scan, switch-scan, and 2D-shift-scan with a shift size of 7) across different linear model. All models are trained from scratch for 100 epochs at a resolution of 224 with a patch size of 16, resulting in a sequence length of 196.}
\begin{tabular}{@{} l S[table-format=2.2]
                   S[table-format=2.2]
                   S[table-format=2.2]
                   S[table-format=2.2] @{}}
\toprule
          & \multicolumn{4}{c}{Top-1 Acc (\%)} \\
\cmidrule(l){2-5} 
Model     & {uni-scan} & {flip-scan} & {switch-scan} & {2d-shift-scan} \\
\midrule
DeltaNet  & 75.75 & 71.22 & 72.63 & 72.16 \\
HGRN      & 74.11 & 71.22 & 70.80 & 69.45 \\ 
RetNet    & 64.03 & 58.06 & 54.48 & 55.57 \\
\bottomrule
\end{tabular}
\label{tab:many_linear_attn_imgnet1k_acc}
\end{table}

\begin{table}[htbp]
\centering
\caption{Top-1 accuracy (\%) on the ImageNet-1K validation set for hybrid DeltaNet models with 30M parameters and 12 layers. For DeltaNet hybrid architectures (denoted as "DeltaNet + ..."), odd-numbered layers utilize DeltaNet, while even-numbered layers employ the specified attention mechanism as token mixers. We use an input resolution of 448 with a patch size of 16, resulting in a sequence length of 784. For 2D sliding window attention, we adopt the implementation from \cite{hassani2023natten}. All models are trained from scratch for 100 epochs.}

\label{tab:hybrid_imgnet1k_acc_784}
\setlength{\tabcolsep}{4pt} 
\begin{tabular}{@{} >{\raggedright\arraybackslash}p{0.70\textwidth} S[table-format=2.2] @{}}
\toprule
Model / Configuration & {Top-1 Acc (\%)} \\
\midrule
\rowcolor{lightblue}
\multicolumn{2}{@{}l}{\textit{Baseline}s} \\
\addlinespace 
Pure DeltaNet & 64.17 \\
Pure 2D SWA (window size=14x14) & 66.83 \\
DeltaNet + Full Attention & 68.72 \\
\midrule
\rowcolor{lightblue}
\multicolumn{2}{@{}l}{\textit{Hybrid with 1D Attention}} \\
\addlinespace
DeltaNet + 1D BA (block size=128) & 66.58 \\
DeltaNet + 1D BA (block size=196) & 66.47 \\
DeltaNet + 1D SWA (window size=128) & 67.19 \\
DeltaNet + 1D SWA (window size=256) & 67.91 \\
\midrule
\rowcolor{lightblue}
\multicolumn{2}{@{}l}{\textit{Hybrid with 2D Attention}} \\
\addlinespace
DeltaNet + 2D BA (block size=14x14) & 63.66 \\
DeltaNet + 2D SWA (window size=14x14) & 68.17 \\
\bottomrule
\end{tabular}
\end{table}

\begin{table}[htbp]
\centering
\caption{Top-1 accuracy (\%) and training time for hybrid DeltaNet models with sparse sequences on ImageNet-1K. For DeltaNet hybrid architectures (denoted as "DeltaNet + ..."), odd-numbered layers utilize DeltaNet, while even-numbered layers employ the specified attention mechanism. Input resolution is 448, patch size is 7 (sequence length 4096). All models are trained from scratch for 100 epochs.}
\label{tab:hybrid_imgnet1k_acc_4k}
\setlength{\tabcolsep}{4pt}
\begin{tabular}{@{} >{\raggedright\arraybackslash}p{0.60\textwidth} c S[table-format=2.2] @{}}
\toprule
Model / Configuration & Training Time & {Top-1 Acc (\%)} \\
\midrule

\rowcolor{lightblue}
\multicolumn{3}{@{}l}{\textit{Baselines}} \\
\addlinespace
DeltaNet + Full Attention & 14h 42m & 66.11 \\
\midrule

\rowcolor{lightblue}
\multicolumn{3}{@{}l}{\textit{Hybrid with 1D Attention}} \\
\addlinespace
DeltaNet + 1D SWA (window size=256) & -- & 62.35 \\
\midrule

\rowcolor{lightblue}
\multicolumn{3}{@{}l}{\textit{Hybrid with 2D Attention}} \\
\addlinespace
DeltaNet + 2D SWA (window size=16x16) & 13h 03m & 66.34 \\
DeltaNet + 2D STA (window size=32x16, tile size=16x8) & -- & 65.28 \\
DeltaNet + 2D STA (window size=48x24, tile size=16x8) & 10h 55m & 66.41 \\
\bottomrule
\end{tabular}
\end{table}

\section{Evaluate Design Choices}

This section evaluates two primary strategies for extending 1D linear models to ND data: scanning and attention-based hybrid architectures on ImageNet classification to guide our architecture design.

\subsection{Scanning or Hybrid?}

We evaluate a range of scanning methods using DeltaNet~\cite{yang2024deltanet} on ImageNet with a short sequence length, with the baseline being Transformer. For these short sequences, full attention is computationally feasible and is thus used in the hybrid model. In Table~\ref{tab:deltanet_imgnet1k_acc}, we compare several scanning methods (primarily single-pass) along with full attention hybrid models. We observe that uni-scan, as a simple baseline, already yields good performance, whereas many single-pass scanning variants degrade performance compared to the baseline. Bi-scan and multi-head-scan fail to show notable improvement over the uni-scan baseline. In contrast, integrating full attention into several layers yields a notable performance boost. Regarding the position and number of attention layers, we find that:

\begin{enumerate}
    \item Integrating attention into half of the layers yields performance comparable to Transformer.
    \item Interleaving attention layers across the model achieves better performance than stacking attention only in the deeper layers, as discussed in~\cite{hatamizadeh2024mambavision}.
    \item In our implementation, scanning alone fails to improve performance and can even be detrimental.
\end{enumerate}

To corroborate this finding, we evaluate single-pass scanning using more linear model types in Table~\ref{tab:many_linear_attn_imgnet1k_acc}. The results consistently show that uni-scan outperforms other single-pass variants, reaffirming that these scanning methods offer no advantages over a simple uni-scan baseline..

Meanwhile, we evaluate cross-scan using a smaller batch size due to its significant memory overhead. As shown in Table~\ref{tab:cross_scan_compare}, we observe that:

\begin{enumerate}
    \item[5.] Although cross-scan outperforms uni-scan, it incurs significantly higher latency and memory overhead. Moreover, its performance still lags behind that of the Transformer.
    \item[6.] Integrating just two full attention layers achieves even higher accuracy, making the cost of cross-scan unjustified.
\end{enumerate}

\textbf{In summary, scanning methods offer little to no performance benefit compared to forming a hybrid model with attention.} Therefore, hybrid attention is a more promising choice. However, when processing longer sequences, even several full attention layers can impose significant latency, making local attention a necessary component for processing longer sequences efficiently.

\subsection{Choosing the Right Local Attention}

As noted in Section~\ref{local_attn_def}, there are two types of local attention: Block Attention, which operates in non-overlapping blocks, and high-order SWA with a flexible local window. We evaluate their effectiveness using a longer sequence length (784) with DeltaNet on ImageNet. The results are presented in Table~\ref{tab:hybrid_imgnet1k_acc_784}. The baselines are pure DeltaNet with no scanning, pure 2D SWA, and DeltaNet + full attention. For all the local attentions, we consider both 1D and 2D variants, and we use NATTEN~\cite{hassani2023natten} as the implementation for 2D SWA.

From the results, we can conclude that:

\begin{enumerate}
    \item The choice of local attention must account for data dimensionality, as 2D local attention consistently outperforms its 1D counterpart.
    \item High-order SWA outperforms Block Attention in both 1D and 2D scenarios. This is likely due to the restrictive boundary effects imposed by Block Attention's non-overlapping partitions.
    \item A 2D SWA hybrid achieves performance comparable to a full attention hybrid.
    \item Both pure DeltaNet and pure 2D SWA underperform compared to their hybrid counterpart.

\end{enumerate}

While vanilla high-order SWA performs well, it offers no significant speedup over full attention, despite the high sparsity. This limitation motivates our adoption of Sliding Tile Attention (STA~\cite{zhang2025sta_attn}), a hardware-efficient variant of high-order SWA.

\subsection{Sliding the Window by Tiles Instead of Tokens}

STA is a hardware-efficient variant of high-order SWA. Rather than shifting the window token-by-token, STA shifts the window tile-by-tile, with tokens within a tile sharing the same window. This way, by ensuring that the number of tokens within a tile equals the block size used in Flex Attention~\cite{dong2024flex}, no mixed blocks are generated, which enables tangible hardware speedups over full attention. \footnote{The official STA implementation primarily uses kernels written in ThunderKittens. For simplicity, we implement STA using Flex Attention while preserving the same high-level design.}.

We evaluate the effectiveness of STA in Table~\ref{tab:hybrid_imgnet1k_acc_4k}. Since the block size used in Flex Attention is required to be at least 128, we use a longer sequence length of 4096. The baseline is the full attention hybrid. From the table, we can conclude that:

\begin{enumerate}
    \item As previously observed, dimensionality is critical, with 2D SWA variants significantly outperforming their 1D counterparts.
    \item The STA hybrid achieves performance comparable to the full attention hybrid with faster speed. While 2D SWA ensures performance, it provides no notable speedup over full attention.
\end{enumerate}

Therefore, for the integration of local attention in processing long sequences, STA stands out as both a fast and expressive local attention mechanism compared to vanilla high-order SWA and Block Attention. We note that our experiments are limited to moderately long sequences due to the computational cost of including a full attention baseline.

\section{Efficient N-Dimensional Attention}

\subsection{Architecture Overview}

We name our proposed architecture, a hybrid of linear recurrence and efficient high-order SWA, Efficient N-dimensional Attention (ENA). In ENA, the window size of SWA is configurable, with full attention being a special case corresponding to a large window. A simple illustration is shown in Fig.~\ref{fig:arc}. We highlight several key aspects of ENA below:

\begin{enumerate}
\item Unlike many prior works that focus on domain-specific enhancements to linear models, such as introducing multi-scale processing, modifying gating mechanisms, or altering update rules, this work directly adopts the linear model implementations from FLA~\cite{yang2024fla}, originally designed for language tasks. These domain-specific enhancements are orthogonal to ENA's architecture. Notably, these linear models, originally designed for language, already yield competitive results on vision tasks without domain-specific modifications.
\item ENA employs a simple block-wise hybrid design for ease of implementation and simplicity. However, when adapting to other established architectures, the linear model and local attention components can also be placed within the same layer.
\item We do not focus on scanning, as our experiments show its benefits are marginal compared to those of a hybrid architecture. Unless otherwise specified, we use uni-scan without any permutation to the sequence, even when using a causal linear model.
\item We treat full attention as a special case of SWA (with a large window) and use it for short sequences (<1K tokens) where local attention offers no efficiency benefits. As shown in Section~\ref{sparsity}, ENA with full attention (i.e., half linear recurrence and half attention) sometimes even outperforms the Transformer (which uses full attention throughout all the layers).
\item We primarily use DeltaNet as the linear recurrence module in ENA due to its strong empirical performance. Other types of linear model are discussed in previous section (e.g., HGRN, RetNet) and we aim to include more comprehensive discussion in future.
\item Throughout this paper, \texttt{ENA} denotes our proposed hybrid architecture. We specify the attention type, e.g., \texttt{ENA-DeltaNet-Full-Attention}, to distinguish between full attention and STA variants.

\begin{table}[htbp]
\centering
\caption{Top-1 accuracy (\%) on the K400 test set for hybrid DeltaNet models with sparse sequences. For hybrid architectures (denoted as "DeltaNet + ..."), odd-numbered layers use DeltaNet, while even-numbered layers employ the specified attention mechanism as token mixers. We evaluate models at input resolution 224 with patch sizes of 16 and 14, using 32 frames. Since the sequence length is long enough, ENA uses 3D STA, and the corresponding window size and tile size are provided in the table. The resulting sequence lengths are 6272 and 8192, respectively. All models are initialized from distilled image models and trained for 25 and 20 epochs.}
\label{tab:hybrid_k400_acc}
\setlength{\tabcolsep}{4pt}
\begin{tabular}{@{} >{\raggedright\arraybackslash}p{0.70\textwidth} S[table-format=2.2] @{}}
\toprule
Model / Configuration & {Top-1 Acc (\%)} \\
\midrule

\rowcolor{lightblue}
\multicolumn{2}{@{}l}{\textit{Sequence Length = 6272}} \\
\addlinespace
Pure DeltaNet & 69.74 \\
ENA-DeltaNet-Full Attention & 75.14 \\
ENA-DeltaNet-STA (window size=32×6×6, tile size=32×2×2) & 73.79 \\
\midrule

\rowcolor{lightblue}
\multicolumn{2}{@{}l}{\textit{Sequence Length = 8192}} \\
\addlinespace
Pure STA (window size=24×12×12, tile size=8×4×4) & 65.90 \\
ENA-DeltaNet-Full Attention & 72.28 \\
ENA-DeltaNet-STA (window size=24×12×12, tile size=8×4×4) & 72.63 \\
\bottomrule
\end{tabular}
\end{table}

\end{enumerate}

While combining linear models with SWA is not new, as explored in~\cite{arora2024based}, ENA can be viewed as a natural extension of this idea, with the key distinction being the N-dimensional data. We additionally conduct several experiments to validate its performance.

\subsection{ENA in 3D Understanding}

To validate the architectural principles established in 2D image classification, we now evaluate ENA's performance on 3D video classification. For datasets, we use K400~\cite{kay2017k400}. The video model is initialized using image classification models distilled from a ImageNet-1K finetuned version of SigLIP2. We evaluate two settings: sequence lengths of 6K and 8K, trained for 25 and 20 epochs, respectively. Given the long sequence length, ENA employs 3D STA in these experiments. The results are shown in Table~\ref{tab:hybrid_k400_acc}, from which we can conclude that:

\begin{enumerate}
    \item STA steadily improves pure linear recurrence even in 3D scenarios. As shown in the 6K and 8K settings, adding STA improves the performance notably.
    \item For high-order data, STA must leverage locality across all dimensions to maintain high performance. In the 6K setting, we do not utilize the locality of the time dimension, thus resulting in worse performance than full attention hybrids. However, in the 8K setting, we utilize all three dimensions' locality, resulting in even better performance than the full attention hybrid.
    \item The pure STA model in the 8K setting serves as an ablation for the effectiveness of linear recurrence. The inferior performance of the pure STA model confirms the necessity of the linear recurrence component in our hybrid architecture.
\end{enumerate}

\begin{table}[h]
\centering
\begin{tabular}{lccccc}
\toprule
Iter & FID$\downarrow$ & sFID$\downarrow$ & IS$\uparrow$ & Pre.$\uparrow$ & Rec.$\uparrow$ \\
\midrule
400k     & 4.7 & 4.7 & 170.1 & 0.83 & 0.58 \\
600k  & 4.4 & 4.6 & 174.1 & 0.83 & 0.60 \\
600k (mh2d-scan)  & 4.6 & 4.8 & 170.7 & 0.82 & 0.59 \\
\bottomrule
\end{tabular}
\caption{Quantitative generative performance metrics of \textit{ena-deltanet-xl-gen2d} on ImageNet 512 $\times$ 512 over 50k samples.}
\label{tab:gen2d_tab}
\end{table}

\begin{table}[ht]
\centering
\begin{tabular}{lccccc}
\toprule
Models& FID$\downarrow$ & sFID$\downarrow$ & IS$\uparrow$ & Pre.$\uparrow$ & Rec.$\uparrow$ \\
\midrule
SiT& 5.88 & 4.98 & 150.51 & 0.84 & 0.56 \\
ENA-Deltanet-Full Attention& 5.02 & 5.16 & 165.71 & 0.82 & 0.61 \\
ENA-DeltaNet-STA (window size=24×24, tile size=8×8)& 4.87 & 4.81 & 165.21 & 0.83 & 0.58 \\
\bottomrule
\end{tabular}
\caption{Quantitative generative performance metrics for different models on ImageNet 512 $\times$ 512 over 50k samples. All models were trained for 400k iterations with a learning rate decaying from $1\times 10^{-3}$ to $1\times 10^{-4}$ after a 10k warmup period.}
\label{tab:gen2d_tab_2}
\end{table}

\subsection{ENA in 2D Generation}

We train an image generation model using REPA~\cite{yu2024repa}. The model builds on the SiT~\cite{ma2024sit} architecture, using DeltaNet as the linear model and STA as the local attention. We use half the batch size of the original REPA setting and train for 400K and 600K steps (equivalent to 200K and 300K steps in REPA's setting, respectively). We first train the model for 400K steps, then fine-tune it for an additional 200K steps using both uni-scan and multi-head-2d-scan to compare their effects. We use Muon as the optimizer and adopt a higher overall learning rate for faster convergence.

Since the maximum sequence length of REPA is 1024 when using a resolution of $512 \times 512$ (due to feature alignment constraints from pretrained encoders), we apply a window size of $24 \times 24$ in STA only, as a proof-of-concept. This leads to only $44\%$ sparsity and is not optimal in terms of efficiency, as the minimum block size of Flex Attention should be 128, while we use 64 here. We train a model with roughly 366M parameters, referred to as \textit{ena-deltanet-sta-w24x24-t8x8-xl-gen2d}.

Selected qualitative generation results are shown in Fig.~\ref{fig:gen2d}, and quantitative metrics are provided in Table~\ref{tab:gen2d_tab}. We report Frechet Inception Distance (FID$\downarrow$), sFID$\downarrow$, Inception Score (IS$\uparrow$), precision (Pre.$\uparrow$), and recall (Rec.$\uparrow$) using 50K generated samples. Lower values indicate better performance for FID and sFID, while higher values are preferred for IS, precision, and recall. We observe fast convergence at 400K steps (equivalent to 200K steps in REPA), reaching a FID of 4.7. To further evaluate multi-head scanning, we finetune the 400K checkpoint (which uses uni-scan, i.e., no permutation) using both uni-scan and multi-head-2d-scan (denoted as mh2d-scan) for an additional 200K steps. For the mh2d-scan variant, we use a learning rate twice as high as the uni-scan counterpart. This result reinforces our earlier finding that scanning methods have a negligible impact on performance.

To compare against the Transformer and ENA with full attention, we conduct an additional experiment for 400K steps, using a learning rate that decays from $1\times 10^{-3}$ to $1\times 10^{-4}$ after a 10K-step warmup period, with the results shown in Table~\ref{tab:gen2d_tab_2}. From the table, we observe that ENA with full attention generally achieves comparable or even better performance than the Transformer. Meanwhile, ENA with $50\%$-sparsity STA also performs on par with ENA with full attention. These observations confirm our central finding: combining linear recurrence and attention is beneficial, and that further speedups can be obtained by controlling the attention sparsity level.

\begin{figure}[t]
    \centering
    \vspace{-0pt}
    \includegraphics[width=0.99\linewidth]{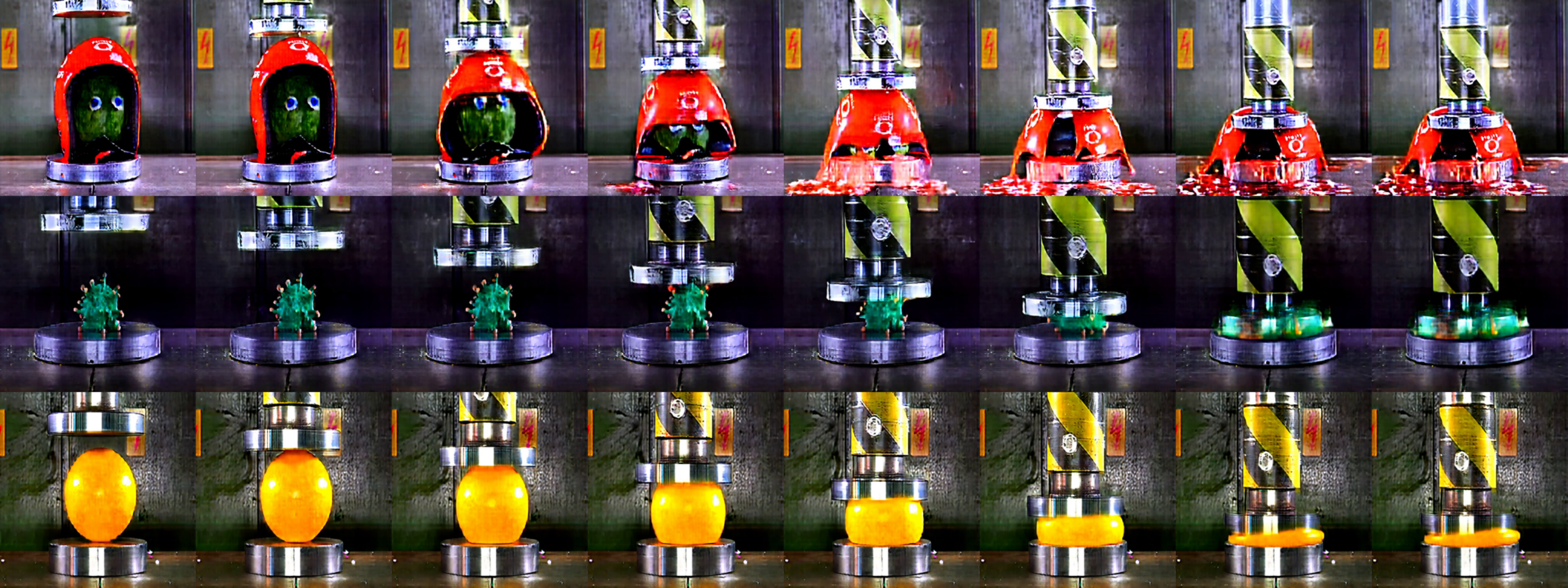}
    \caption{Video generation results on a toy dataset from \textit{ena-deltanet-sta-w3x24x48-t1x8x16-cogvideox-2b}, demonstrating relatively consistent temporal behavior. Due to the limited training steps and simplified training settings, some artifacts and blurriness are observed in the spatial dimensions.
}
    \label{fig:video}
\end{figure}

\subsection{ENA in 3D Generation}

We also briefly explore video generation using ENA. Specifically, we adapt the architecture introduced in CogVideo~\citep{hong2022cogvideo, yang2024cogvideox}, which consists of two attention modules within a single layer: a self-attention and a cross-attention. We replace the self-attention with the linear model used in ENA, and substitute the cross-attention with STA, which supports cross-modality attention mechanisms. Most weights are initialized from the pretrained CogVideoX-2B model, except for newly introduced components such as the short convolution used in the linear model. During training, we freeze all parameters in the channel mixers, updating only the token mixer, thereby minimizing training cost.

As a proof of concept,  we train the model only on a toy dataset of 47 videos. We first adapt the pretrained transformer model to our target resolution of $1024 \times 768$. We then use this adapted transformer as a teacher model. Its noise predictions provide direct supervision for fine-tuning the ENA model. We employ DeltaNet as the linear model and STA with a window size of $3 \times 24 \times 48$ and a tile size of $1 \times 8 \times 16$. Throughout training, we use Muon as the optimizer.

The generation results are shown in Fig.~\ref{fig:video}, exhibiting relatively good temporal consistency. Due to limited computational resources and the high training cost associated with video generation, the model is trained for only around 20K steps, resulting in visible artifacts in individual frames. We leave further scaling and optimization of training settings as future work.

\begin{table}[htbp]
\centering
\caption{The influence of learning rate. Gated linear models perform better in a small learning rate. However, hybrid model (ENA) consistently improve the performance. We use a 12-layer model with a hidden size of 448, resulting in a 30M parameters model. Since training with a sequence length of 196, ENA here uses full attention.}
\label{tab:lr_ablation}
\setlength{\tabcolsep}{5pt}
\begin{tabular}{@{} >{\raggedright\arraybackslash}p{0.55\textwidth} c S[table-format=2.2] @{}}
\toprule
Model / Configuration & Learning Rate & {Top-1 Acc (\%)} \\
\midrule

\rowcolor{lightblue}
\multicolumn{3}{@{}l}{\textit{Sequence Length = 196, 20 Epochs, Warmup Epochs = 20}} \\
\addlinespace
Pure GDP & 2e-3 & 36.34 \\
ENA-GDP-Full Attention & 2e-3 & 55.25 \\
Pure GDP & 5e-4 & 47.29 \\
ENA-GDP-Full Attention & 5e-4 & 49.91 \\
\midrule
Pure GDN & 2e-3 & 34.92 \\
ENA-GDN-Full Attention & 2e-3 & 47.48 \\
Pure GDN & 2e-4 & 34.62 \\
ENA-GDN-Full Attention & 2e-4 & 39.07 \\
\midrule

\rowcolor{lightblue}
\multicolumn{3}{@{}l}{\textit{Sequence Length = 4096, 10 Epochs, Warmup Epochs = 2}} \\
\addlinespace
Pure GDP & 5e-4 & 38.06 \\
ENA-GDP-STA (window size=48x24, tile size=16x8) & 5e-4 & 39.76 \\
\bottomrule
\end{tabular}
\end{table}

\begin{table}[htbp]
\centering
\caption{Performance using Muon as the optimizer. It can be shown that hybrid model (ENA) consistently improves the performance. We use a 12-layer model with a hidden size of 448, resulting in a 30M parameters model. Since training with a sequence length of 4096, ENA here uses STA.}
\label{tab:muon_result}
\setlength{\tabcolsep}{5pt} 
\begin{tabular}{@{} >{\raggedright\arraybackslash}p{0.65\textwidth} S[table-format=2.2] @{}}
\toprule
Model / Configuration & {Top-1 Acc (\%)} \\
\midrule
Pure DeltaNet & 68.95 \\
ENA-DeltaNet-STA (window size=48x24, tile size=16×8) & 74.45 \\
\bottomrule
\end{tabular}
\end{table}

\subsection{Discussions}

\subsubsection{Impact of Learning Rate}

While our main experiments use a learning rate of 2e-3, some gated linear models like Gated DeltaNet (GDN) and Gated DeltaProduct (GDP) are known to prefer smaller learning rates. We conducted experiments in Table \ref{tab:lr_ablation} to verify that ENA's performance gains hold across different learning rates. Since training with a short sequence length, ENA here employs full attention. From the table, we can observe that:

\begin{enumerate}
    \item Smaller learning rates indeed improve the performance of gated linear models.
    \item ENA consistently outperforms pure linear models regardless of the learning rate.
    \item ENA performs better using a larger learning rate than the optimal learning rate for gated linear models, consistent with our default settings.
\end{enumerate}

\subsubsection{Impact of Optimizer}

We use AdamW as the default optimizer. To demonstrate that ENA's benefits are not optimizer-specific, we replaced AdamW with Muon \cite{jordan2024muon} as the optimizer and test the performance in 4K sequence length training with STA. We use a larger learning rate of 4e-3 for Muon. Since training with a long sequence length, ENA here uses 2D STA. From the results shown in Table \ref{tab:muon_result}, we can observe that ENA (the hybrid) performs notably better than the pure DeltaNet model, thus demonstrating that the benefits of ENA are not specific to a single optimizer.

\begin{table}[htbp]
\centering
\caption{Faster training results on ImageNet using Muon. The resolution is 224 with a patch size of 16, resulting in a sequence length of 196. Since training with a short sequence length, ENA here uses full attention.}
\label{tab:speedrun}
\setlength{\tabcolsep}{5pt} 
\begin{tabular}{@{} >{\raggedright\arraybackslash}p{0.40\textwidth} S[table-format=2.2] S[table-format=2.0] S[table-format=2.0] @{}}
\toprule
Model / Configuration & {Top-1 Acc (\%)} & {Epoch} & {Warmup} \\
\midrule
ENA-DeltaNet-Full Attention & 74.27 & 20 & 4 \\
ENA-DeltaNet-Full Attention & 77.08 & 30 & 5 \\
\bottomrule
\end{tabular}
\end{table}

\subsubsection{Training ENA Faster with Muon}

We also find that using Muon results in faster convergence compared to AdamW. We train the tiny-size model (30M parameters) using Muon and a larger learning rate, with a cosine learning rate scheduler. The results are shown in Table \ref{tab:speedrun}, from which we can observe that using Muon can accelerate the convergence speed. This approach achieves performance comparable to 100-epoch training with AdamW in a fraction of the time, making it resource-efficient for future architectural evaluations. Given the observation, we also use Muon as the optimizer for image generation and video generation.

\begin{figure}[th]
    \centering
    \includegraphics[width=0.95\linewidth]{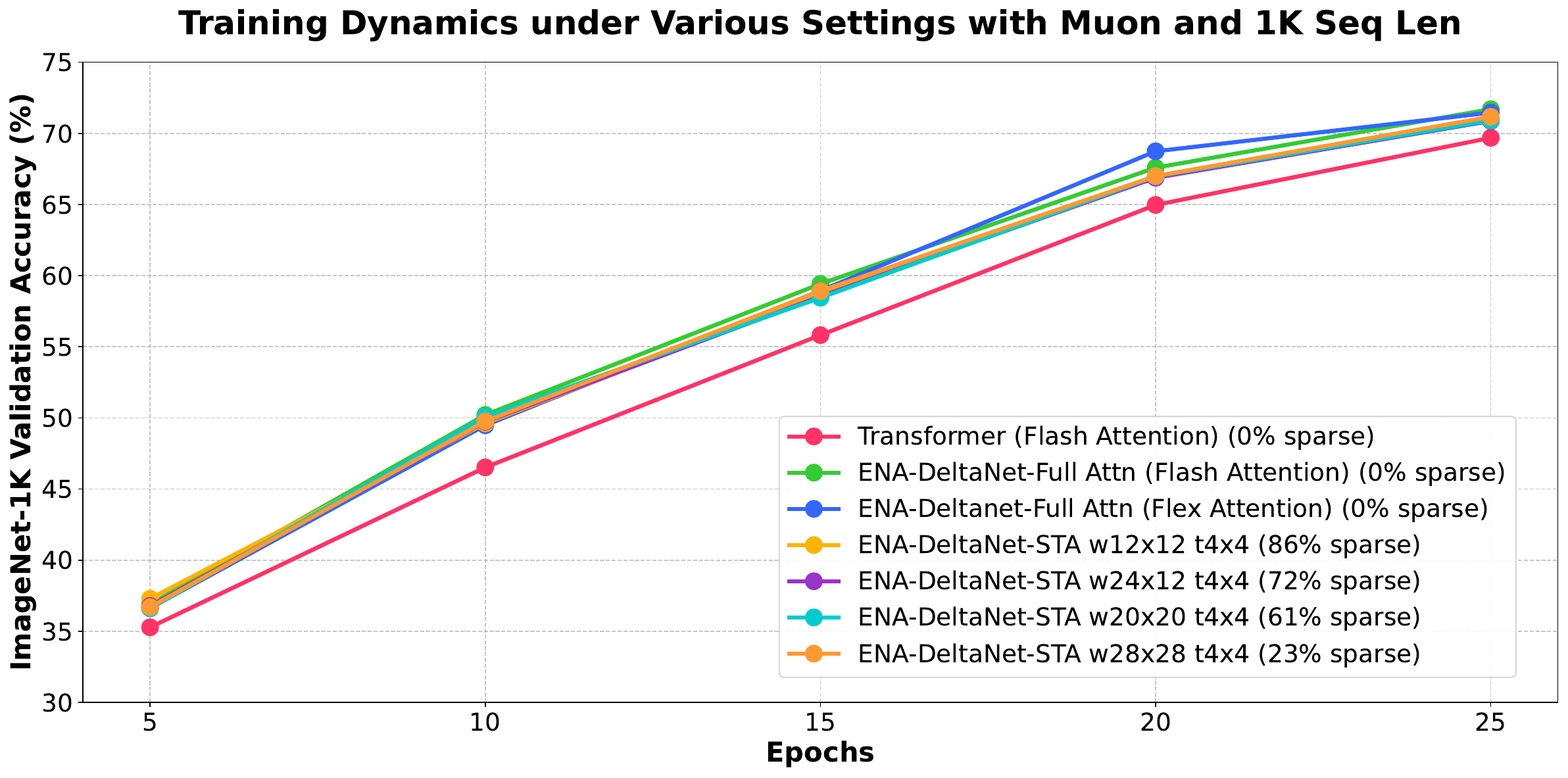}
    \caption{Training dynamics under various settings on ImageNet-1K with Muon and a sequence length of 1024. The hybrid architecture of ENA outperforms the Transformer, and the performance gain from decreasing sparsity levels (i.e., increasing window size) gradually diminishes, indicating diminishing returns. For STA, we use the notation \texttt{w12×12} and \texttt{t4×4} to denote a window size of $12 \times 12$ and a tile size of $4 \times 4$.}
    \label{fig:muon-1k-sparsity}
\end{figure}

\begin{figure}[th]
    \centering
    \includegraphics[width=0.95\linewidth]{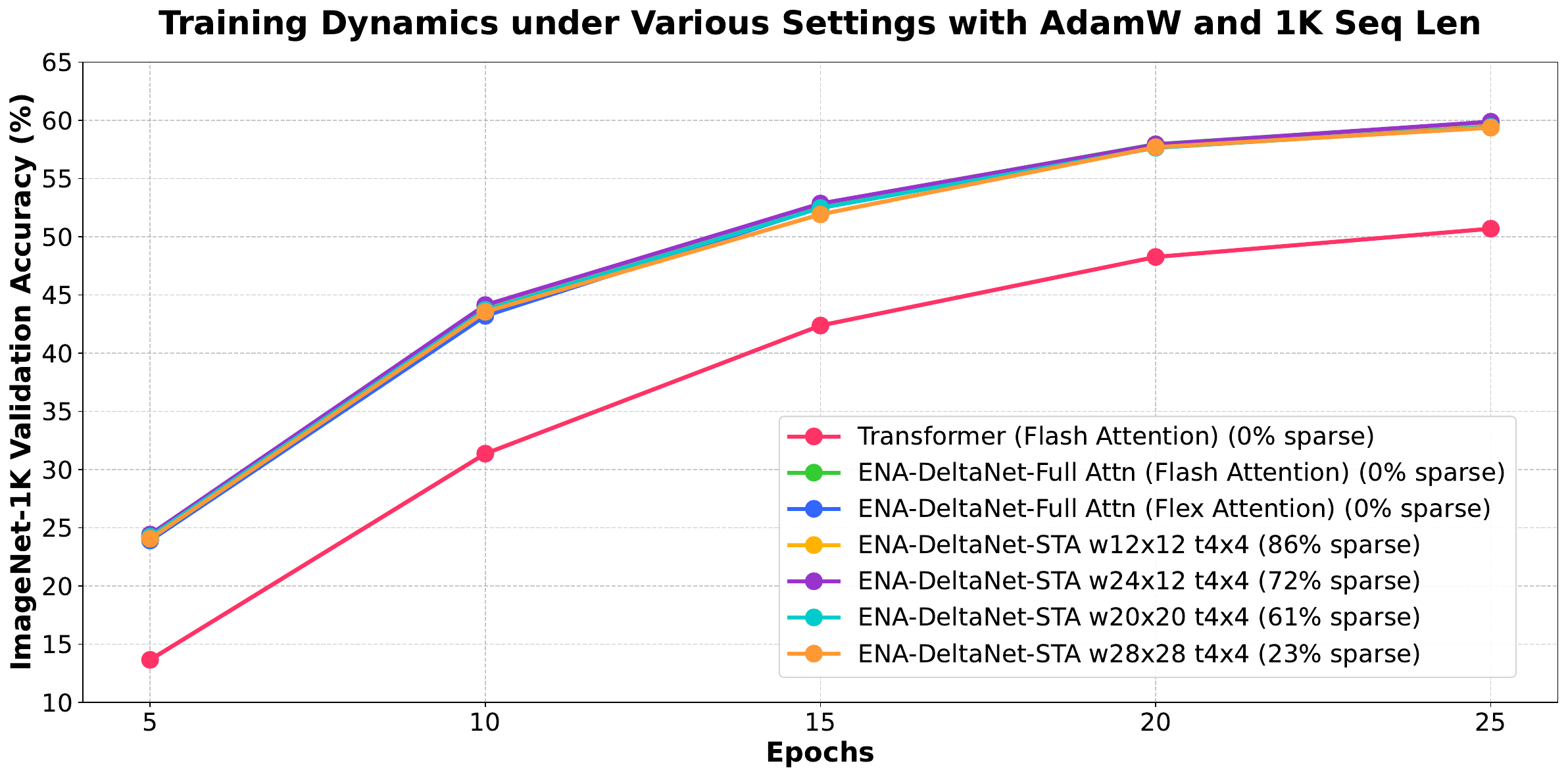}
    \caption{Training dynamics under various settings on ImageNet-1K with AdamW and a sequence length of 1024. Similar conclusions can be drawn from Fig.~\ref{fig:muon-1k-sparsity}. Notably, the performance gap between the Transformer and hybrid models is even larger in this case. For STA, we use the notation \texttt{w12×12} and \texttt{t4×4} to denote a window size of $12 \times 12$ and a tile size of $4 \times 4$.}
    \label{fig:adamw-1k-sparsity}
\end{figure}

\begin{figure}[th]
    \centering
    \includegraphics[width=0.95\linewidth]{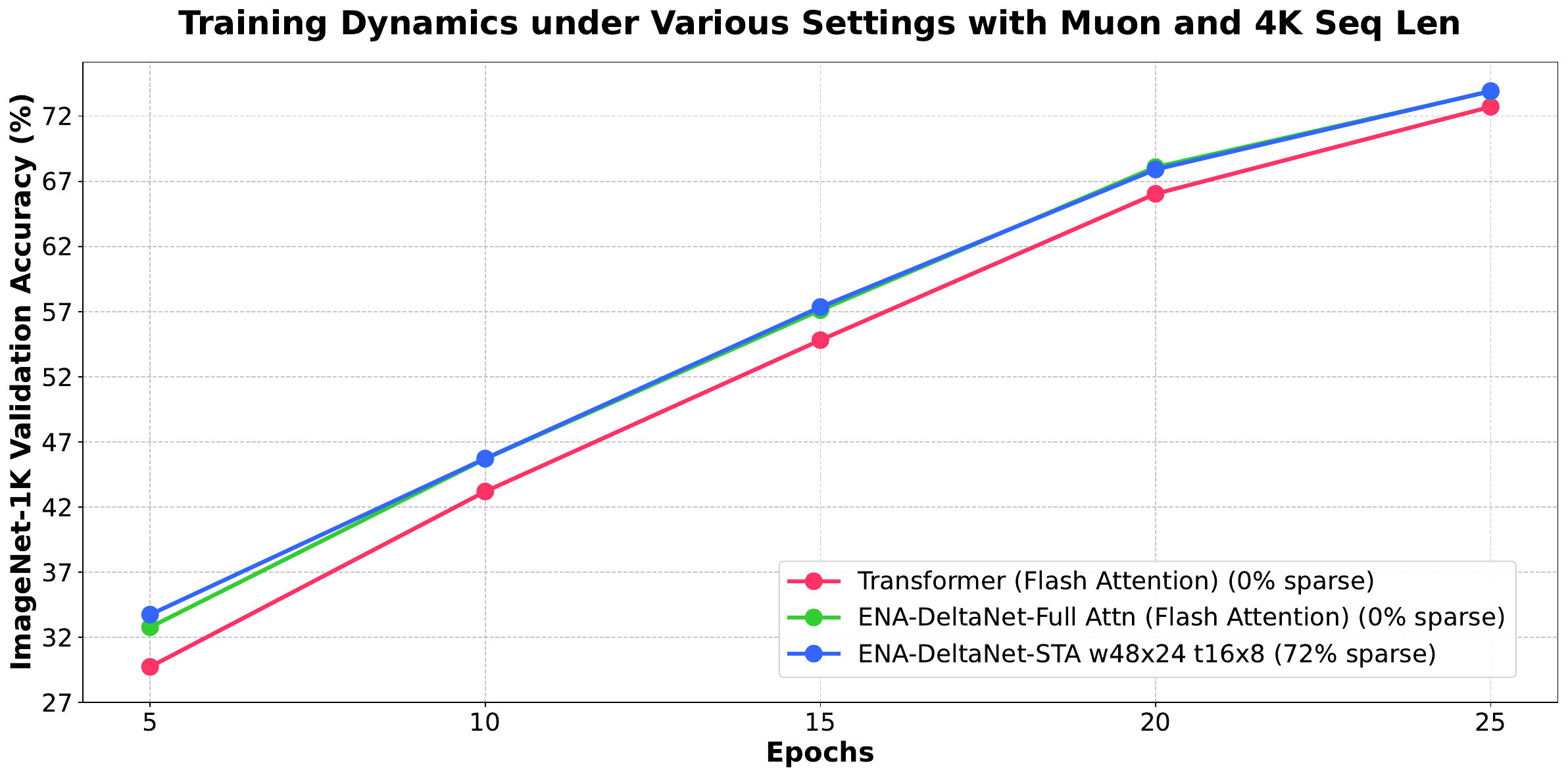}
    \caption{Training dynamics under various settings on ImageNet-1K with Muon and a sequence length of 4096. Similar conclusions can be drawn from Fig.~\ref{fig:muon-1k-sparsity} and Fig. \ref{fig:adamw-1k-sparsity}. Notably, with approximately $70\%$ sparsity, ENA with STA even outperforms ENA using full attention. For STA, we use the notation \texttt{w48×24} and \texttt{t16×8} to denote a window size of $48 \times 24$ and a tile size of $16 \times 8$.}
    \label{fig:muon-4k-sparsity}
\end{figure}

\begin{figure}[th]
    \centering
    \includegraphics[width=0.95\linewidth]{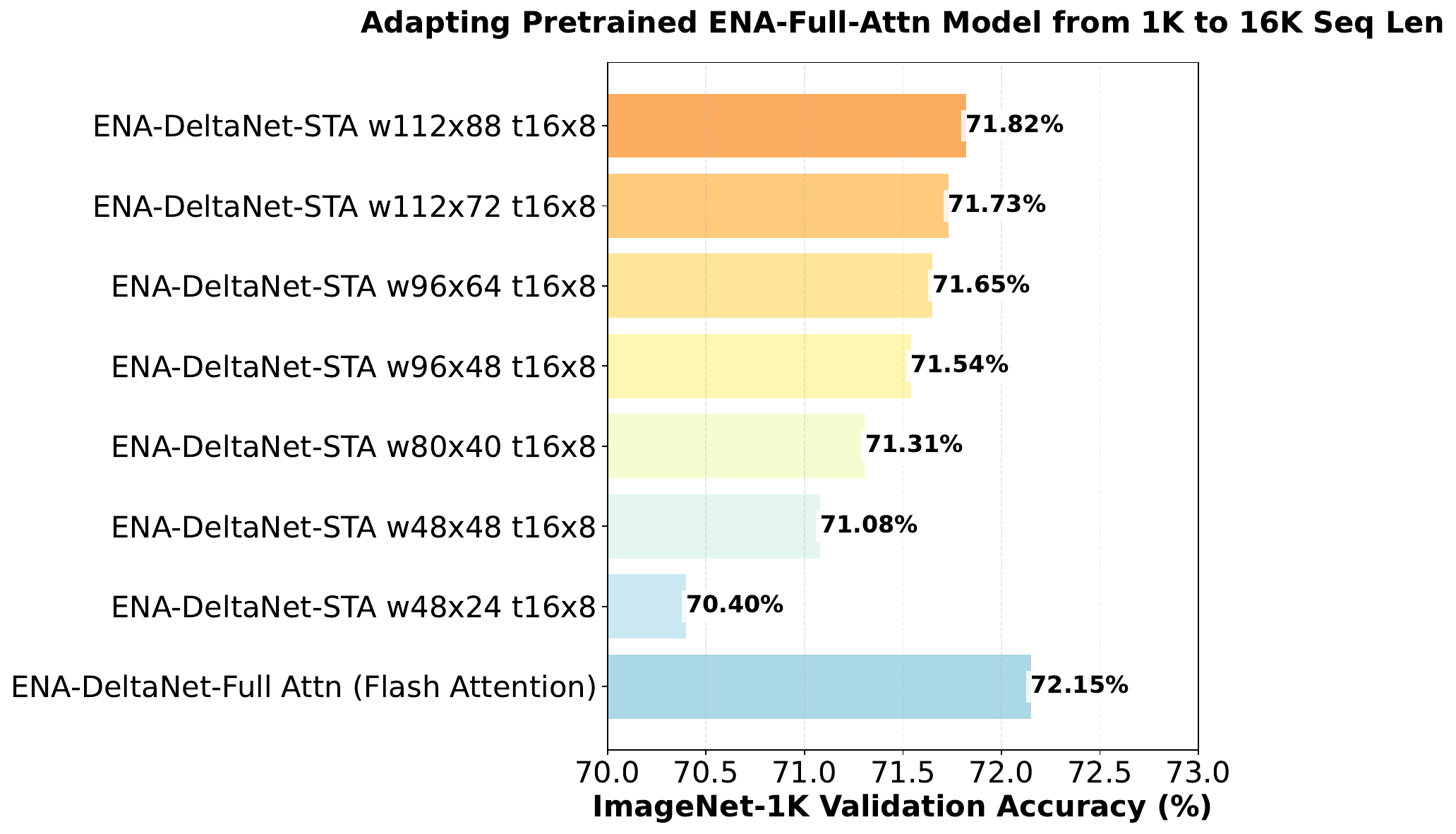}
    \caption{ImageNet-1K validation accuracy at a sequence length of 16,384, obtained by finetuning ENA models pretrained with a sequence length of 1K using full attention under various settings. Reducing sparsity by increasing the window size leads to diminishing performance gains, with a sparsity level of approximately 70\% achieving performance comparable to full attention.}
    \label{fig:muon-16k-sparsity}
\end{figure}

\subsubsection{Impact of Sparsity Levels}
\label{sparsity}

The attention in ENA can have varying sparsity levels, determined by the window size used in local attention. Different levels of sparsity affect both hardware efficiency and model performance. To investigate this, we first conduct a set of experiments on ImageNet-1K pretraining with a sequence length of 1K as a proof of concept. The image size is set to 128 and the patch size to 4. We use both Muon and AdamW optimizers and present the results in Fig.~\ref{fig:muon-1k-sparsity} and Fig.~\ref{fig:adamw-1k-sparsity}, respectively. The batch size is $128 \times 8$, with learning rates of $1\mathrm{e}{-2}$ for Muon and $2\mathrm{e}{-3}$ for AdamW. All models are trained for 25 epochs with 5 epochs of linear warmup. The sparsity level in STA is controlled by the window size with larger windows correspond to lower sparsity. The tile size is fixed at $4 \times 4$.

From the results, we observe that increasing the window size (i.e., reducing sparsity) yields diminishing performance improvements. For example, under the AdamW setting, a window size of $24 \times 12$ (approximately 72\% sparsity) achieves accuracy comparable to full attention (0\% sparsity). Further increasing the window size to $28 \times 28$ does not lead to noticeable performance gains. A similar trend is observed when using Muon, where a window size of $12 \times 12$ already matches the performance of full attention, and larger windows incur additional computation without meaningful improvements.

We present the 4K sequence length training results using AdamW in Table~\ref{tab:hybrid_imgnet1k_acc_4k}, where ENA with a window size of $48 \times 24$ for STA (corresponding to $72\%$ sparsity) matches the performance of ENA with full attention. To support a more robust conclusion, we also report 4K training results using Muon in Fig.~\ref{fig:muon-4k-sparsity}, which confirm the same observation.

To evaluate the generalization of this trend in longer sequences, we finetune the ENA model pretrained with full attention (from Fig.~\ref{fig:muon-1k-sparsity}) on ImageNet-1K at a sequence length of 16,384. The image size is set to 512 and the patch size remains 4. All other parameters are kept the same, except for a randomly initialized learnable positional embedding. We use a constant learning rate of $1\mathrm{e}{-4}$ and train for 2 epochs. As shown in Fig.~\ref{fig:muon-16k-sparsity}, the same observation holds: increasing the window size leads to diminishing gains, with a sparsity level around 70\% once again achieves performance comparable to full attention. It is important to note that this observation can also be found in Table \ref{tab:hybrid_imgnet1k_acc_4k}, where ENA with $70\%$ sparse STA yields performance comparable to ENA with full attention.

Overall, our findings on the impact of sparsity levels in STA can be summarized as follows:

\begin{enumerate}
    \item The strong locality inherent in many datasets means that full attention often performs redundant computations on distant, irrelevant tokens. Local attention leverages this redundancy to improve efficiency while maintaining comparable performance.
    \item Reducing sparsity by increasing the window size yields diminishing, and even negligible, performance gains.
    \item To maintain a consistent trade-off between performance and efficiency across different sequence lengths, fixing the sparsity level is more effective than fixing the window size. As shown in Fig.~\ref{fig:muon-16k-sparsity}, overly high sparsity levels can degrade performance. Based on our experiments, a sparsity level of around 70\% (i.e., each token attends to only 30\% of the sequence) offers a good trade-off.
\end{enumerate}

\subsection{Why a Hybrid Model over Transformer?}
\label{why_hybrid}

This paper mainly introduces a hybrid architecture that combines linear recurrence with high-order SWA, where we count full attention as a special case (in which the window size is large enough to cover all tokens). A natural question is: why prefer a hybrid model over a standard Transformer? We offer several reasons, as partially demonstrated by previous experiments.

\begin{enumerate}
    \item An ENA model with full attention already replaces half of the layers in Transformer with linear recurrence, which has linear time complexity and is much faster when modeling long sequences. This setup already performs comparably to or sometimes better than Transformer, as shown in Table~\ref{tab:deltanet_imgnet1k_acc}, Table \ref{tab:gen2d_tab_2}, Fig.~\ref{fig:muon-1k-sparsity}, and Fig.~\ref{fig:adamw-1k-sparsity}. A more comprehensive evaluation and explanation remains as future work.
    
    \item Furthermore, the full attention component can be replaced with high-order SWA to achieve greater speedups with minimal performance loss, as shown in previous sections. By carefully controlling the sparsity level in SWA, we can achieve similar performance than using full attention.
\end{enumerate}

To conclude, simply combining half linear recurrence and half full attention yields comparable performance to Transformer, and the full attention part can be further accelerated using high-order SWA with a high sparsity level, without notable performance degradation.

\subsection{Distillation from Pretrained Models}

\label{distill}

Distilling from pretrained models is a resource-efficient alternative to training from scratch. To evaluate ENA's effectiveness as a student model, we distill knowledge from SigLIP2-Base~\cite{tschannen2025siglip2}. \textbf{For our base-sized model, we initialize the majority of weights directly from the teacher model.} For smaller models, we randomly initialize the weights. We train on ImageNet-1K using temperature-scaled KL divergence between student and teacher outputs for 200 epochs, followed by 10 epochs of supervised finetuning. The results of several models on ImageNet-1K are shown in Table. \ref{tab:distill}. For naming consistency, we adopt the template \texttt{ena-\{linear model type\}-\{attention type\}-\{size\}-\{task\}-\{teacher info\}} throughout this paper.

\begin{table}[htbp]
\centering
\caption{Performance of distilled model on ImageNet-1K validation set.}
\label{tab:distill}
\setlength{\tabcolsep}{5pt} 
\begin{tabular}{@{} >{\raggedright\arraybackslash}p{0.55\textwidth} S[table-format=2.2] @{}}
\toprule
Models & {Top-1 Acc (\%)} \\
\midrule
ena-deltanet-fullattn-base-imgc-siglip2-base-p16-224 & 85.15 \\
ena-lact-fullattn-base-imgc-siglip2-base-p16-224 & 84.95 \\
ena-hgrn-fullattn-base-imgc-siglip2-base-p16-224 & 84.16 \\
\bottomrule
\end{tabular}
\end{table}

\subsection{Hardware Efficiency}

We compare the hardware efficiency of ENA against a Flash Attention (FA) based Transformer. Both ENA and FA models use a 12-layer encoder with a hidden size of 224, and we report the results in Fig.~\ref{fig:speed}. For linear recurrence, we use DeltaNet implemented in FLA \cite{yang2024fla} with uni-directional scan. We use full attention for short sequences and switch to Flex Attention implementation of STA when the sequence length exceeds 1152. During training, we apply an MSE loss on the hidden states and use a window size that covers $30\%$ of the tokens for STA. From the figure, we observe that:

\begin{enumerate}
    \item ENA's training and inference times scale more favorably than FA's, offering notable speedups for sequence lengths in the thousands.
    \item Although ENA's memory consumption is slightly higher than FA's, the difference is minor and can potentially be reduced through kernel fusion. It is also worth noting that both the linear model and STA in our implementation are built using Triton, while FA (specifically, FA2) is implemented directly in CUDA with extensive optimizations for A100 GPUs. As demonstrated in the original STA paper, further efficiency gains may be achieved by developing specialized kernels using CUDA or ThunderKittens~\cite{spector2024thunderkittens}. This remains a promising direction for future work.
\end{enumerate}

In conclusion, ENA provides a compelling alternative to the Transformer for long-sequence modeling, achieving notable speedups with comparable memory usage.

\section{Related Works}

Linear recurrent models are token mixers with states and linear time complexity. Representative ones include RetNet~\citet{sun2023retnet}, HGRN~\citet{qin2024hgrn2}, GLA~\citet{yang2024gla}, GSA~\cite{zhang2024gsa}, RWKV~\citet{peng2023rwkv}, DeltaNet~\citet{yang2024deltanet}, Gated DeltaNet~\citet{yang2024gdn}, RWKV-7~\citet{peng2025rwkv7}, LaCT~\cite{zhang2025tttdoneright}, and MesaNet~\cite{vonoswald2025mesanet}. Other sub-quadratic token mixers, such as Log-Linear Attention \cite{guo2025loglinearattention} have also been proposed to balance efficiency and expressiveness. Although ENA is built on linear recurrence, its hybrid framework is potentially compatible with any sub-quadratic token mixer.

Prior works have also utilized local attention with linear recurrence, including MambaVision~\cite{hatamizadeh2024mambavision}, TTT-MLP~\cite{dalal2025onemin-video}, and LaCT~\cite{zhang2025tttdoneright}. The local attention in these works typically consists of non-overlapping blocks, denoted in this paper as \textit{Block Attention}. Block Attention is simple to implement using Flash Attention but imposes strong restrictions on border tokens. 

High-order SWA, introduced in Neighborhood Attention~\cite{hassani2023natten} extends SWA from language modeling and, unlike Block Attention, imposes no border restrictions. However, naive implementation of high-order SWA generates many mixed blocks, which prevents actual speedup despite sparsity and reduced FLOPs. 

Sliding Tile Attention (STA \cite{zhang2025sta_attn}) addresses this by sliding the window by tiles, where a number of tokens share the same window. In this way, by ensuring that the number of tokens within a tile equals the block size used in Flex Attention \cite{dong2024flex}, no mixed blocks are generated, thus achieving tangible hardware speedups.

\section{Conclusion}

In this paper, we evaluated two primary strategies for adapting linear recurrent models to high-order data: scanning and attention integration. Our findings show that while scanning offers negligible benefits, integrating efficient high-order SWA yields notable performance gains. Finally, we propose ENA, a simple hybrid of linear recurrent models and high-order SWA. ENA is a general architecture applicable to any linear model, and in this work, we primarily demonstrate its effectiveness using DeltaNet.

\section{Acknowledgments}

We thank Xufang Luo from Microsoft Research for her support and insightful discussions, which made this work possible. We thank Songlin Yang and Yu Zhang for their excellent kernel implementations in the Flash Linear Attention (FLA) repository. We also thank the authors of STA for recognizing the hardware inefficiency of vanilla high-order SWA and proposing a simple yet effective solution.

\clearpage

\bibliography{iclr2025_conference}
\bibliographystyle{iclr2025_conference}

\end{document}